\documentclass[conference]{IEEEtran}
\IEEEoverridecommandlockouts
\usepackage{cite}
\usepackage{amsmath,amssymb,amsfonts}
\usepackage{algorithmic}
\usepackage{graphicx}
\usepackage{textcomp}
\usepackage{booktabs}
\usepackage{xcolor}
\usepackage{comment}
\usepackage{bm}
\usepackage{float} 
\def\BibTeX{{\rm B\kern-.05em{\sc i\kern-.025em b}\kern-.08em
    T\kern-.1667em\lower.7ex\hbox{E}\kern-.125emX}}

\title{Exploring Lexical Irregularities in Hypothesis-Only Models of Natural Language Inference\\
}

\makeatletter
\newcommand{\linebreakand}{
  \end{@IEEEauthorhalign}
  \hfill\mbox{}\par
  \mbox{}\hfill\begin{@IEEEauthorhalign}
}
\makeatother

\author{
\IEEEauthorblockN{Qingyuan Hu}
\IEEEauthorblockA{\textit{Computer and Information Technology} \\
\textit{Purdue University}\\
West Lafayette, IN, USA \\
hu528@purdue.edu}
\and
\IEEEauthorblockN{Yi Zhang}
\IEEEauthorblockA{\textit{Computer and Information Technology} \\
\textit{Purdue University}\\
West Lafayette, IN, USA \\
zhan3050@purdue.edu}
\and
\IEEEauthorblockN{Kanishka Misra}
\IEEEauthorblockA{\textit{Computer and Information Technology} \\
\textit{Purdue University}\\
West Lafayette, IN, USA \\
kmisra@purdue.edu}
\linebreakand
\IEEEauthorblockN{Julia Taylor Rayz}
\IEEEauthorblockA{\textit{Computer and Information Technology} \\
\textit{Purdue University}\\
West Lafayette, IN, USA \\
jtaylor1@purdue.edu}}

\begin{document}

\maketitle

\begin{abstract}
Natural Language Inference (NLI) or Recognizing Textual Entailment (RTE) is the task of predicting the entailment relation between a pair of sentences (premise and hypothesis). This task has been described as “a valuable testing ground for the development of semantic representations”\cite{bowman_large_2015}, and is a key component in natural language understanding evaluation benchmarks. Models that understand entailment should encode both, the premise and the hypothesis. However, experiments by Poliak et al.\cite{poliak_hypothesis_2018} revealed a strong preference of these models towards patterns observed only in the hypothesis, based on a 10 dataset comparison.  Their results indicated the existence of statistical irregularities present in the hypothesis that bias the model into performing competitively with the state of the art. While recast datasets provide large scale generation of NLI instances due to minimal human intervention, the papers that generate them do not provide fine-grained analysis of the potential statistical patterns that can bias NLI models. In this work, we analyze hypothesis-only models trained on one of the recast datasets provided in Poliak et al.\cite{poliak_hypothesis_2018} for word-level patterns. Our results indicate the existence of potential lexical biases that could contribute to inflating the models’ performance.
\end{abstract}


\section{Introduction}

Advancements in Natural Language Processing (NLP) allow for more natural communication between people and computational devices. This natural communication brings a list of desired tasks and requirements to understand and test. One of such desired tasks is an ability to computationally infer information, in a way that is similar to human inference. In this paper we explore a narrow subset of inferences that are used in natural language, namely entailment and contradiction.   

\subsection{Natural Language Inference}
Natural Language Inference (NLI) in general, and its subset, Recognizing Textual Entailment (RTE) in particular, is the task of predicting the entailment relation (using three labels: entailment, neutral, contradiction) between a pair of sentences. The first sentence in the pair is usually referred to as premise, and the second sentence is referred to as hypothesis, see \textsc{table}~\ref{tab:examples} for examples. In an RTE task, given a text, T, and a hypothesis, H: ``T entails H if, typically, a human reading 
T would infer that 
H is most likely true'' \cite{dagan_pascal_2006,poliak_hypothesis_2018}.
This task has been described as ``a valuable testing ground for the development of semantic representations'' \cite{bowman_large_2015}, and is a key component in natural language understanding evaluation benchmarks. Many studies have focused on the modelling of NLI label prediction using different datasets, for example, human elicited SNLI dataset \cite{bowman_large_2015}, human judged SciTail dataset \cite{khot_scitail_2018} and recast dataset SPR\cite{white-etal-2017-inference}. Models that perform well are considered to have achieved a level of language understanding. But do they, really? 

\begin{table}[H]
\centering
\caption{Example of an RTE/NLI task instance}
\label{tab:examples}
\begin{tabular}{|c|c|c|}
\hline
\textbf{Premise} & \textbf{Hypothesis} & \textbf{Label} \\ \hline
\textit{``The brown cat ran"} & \textit{``The animal moved"} & \textsc{entailment} \\ \hline
\begin{tabular}[c]{@{}l@{}}\textit{``A woman is reading}\\ \textit{with a child"}\end{tabular} & \textit{``A woman is sleeping"} & \textsc{contradiction} \\ \hline
\end{tabular}
\end{table}

\subsection{Motivation}
Matching human level of inference by a computational device is not an easy task, and determining how to measure the level or degree of inference is difficult. Yet, it is essential for the task of explainable AI or any system that can claim any level of `knowledge' or reasoning ability, which underlies natural language understanding and human-like communication. Consider the following sentences: \textit{A cat ran towards a tree. When the animal moved, the birds sitting on the tree flew away}. A human usually has no problem understanding that the \textit{animal}  refers to the \textit{cat} in the previous sentence, that \textit{moves} refers to the act of \textit{running}, and, finally, that \textit{animal moving} is inferred from the \textit{running cat}. Moreover, \textit{running cat} and \textit{moving animal} refer to exactly the same event with exactly the same participant of this event. A well-functioning natural language system has to exhibit the same level of understanding in detecting the same events that may be talked about using different words, and detect contradictions allowing it to conclude that it is not the same event. One way to test such understanding is by introducing a pair of sentences, such as about the \textit{running cat} and \textit{moving animal}, and asking for the system to identify relation(ship) between these two sentences.

For the task of such relation identification, an ideal language understanding system should encode both premise and hypothesis to infer the relation label. However, while some models achieved high accuracy, it is questionable whether those models are merely picking up statistical irregularities in the datasets instead of ``understanding'' natural language or the actual logical relationship between the two sentences. In other words, a model that has a high accuracy by predicting irregularities may end up showing a low performance in another NLI task, as the featured irregularities may not be included in every testing set. If a model is trained with a dataset that contains many irregularities, this model cannot be generalized to other datasets since it doesn't pick up background knowledge of real world but patterns of irregularities. More importantly, a model that learns statistical irregularities is not a reliable real world model (which would be indicated by its lack of generalization to other datasets).  

To demonstrate issues with models that perform well on the task without understanding its meaning, Poliak et al. \cite{poliak_hypothesis_2018} proposed a model that encoded only the hypothesis and ignored the premise to infer the entailment relation between pairs of sentences. They referred to this as a ``hypothesis baseline"\footnote{A baseline is a method to create the simplest possible predictions in machine learning. It can be used to evaluate the performance (e.g. accuracy) of a dataset. In this case, the prediction is based only on hypothesis.}. With this baseline being introduced to diagnose datasets, the authors found that the existence of statistical irregularities present in the hypothesis could bias the model into performing competitively with the state of the art performance. For example, the occurrence of negation words such as \textit{nobody} was found to be highly indicative of "contradiction" on the SNLI dataset. Poliak et al.'s
analysis suggests that (a) there are potential irregularities in the datasets that shouldn't affect semantic inference, and (b) future models should outperform the hypothesis-only baseline rather than the majority baseline\footnote{The majority baseline is determined by using the majority class as the default classification.}  to show competence in understanding language.

In this work, we augment results shown by Poliak et al. \cite{poliak_hypothesis_2018} and analyze hypothesis-only models trained on a specific recast (i.e., with minimal human intervention) dataset, namely the Semantic Proto-Roles dataset, provided in \cite{white-etal-2017-inference}, for word-level patterns. The purpose of this work is to explore additional lexical biases that could contribute to inflating the models' performance on NLI tasks, thus potentially being ``right for the wrong reasons" \cite{mccoy2019right}. Our goal here is not to show state-of-the-art performance, but rather highlight how lexical heuristics that have nothing to do with semantic inference can be picked up by models trained on datasets that do not control for them.

\section{Related Work}
It is tempting to talk about natural language inference in terms of sentence understanding and truth of a sentence. If one were to do that, work going back to 1970s \cite{katz1972}, if not earlier, would have to be reviewed. The discussion could start with lexical semantics and include quantification, entailment and subsumption, ambiguity, common sense, and many more (see \cite{bowman-zhu-2019-deep} for an overview). We could talk about the truth of sentences, such as Bertrand Russell's \textit{The present king of France is bald} -- what exactly is its truth value and does it mean when there is no kings in France, presently? One could follow the line of arguments in formal semantics and explore sentence understanding as a set of of conditions or situations under which the sentence is true. Arguably, however, it is much harder than what is intended by NLI. As Wang and Jiang \cite{wang2015learning} pointed out, ``existing solutions to NLI range from shallow approaches based on lexical similarities \cite{glickman2005web} to advanced methods that consider syntax \cite{Mehdad2009SemKerSK}, perform explicit sentence alignment \cite{maccartney2008phrase} or use formal logic \cite{Clark2009AnIA}." However, none of these are concerned with the factual information, only to what is stated in the sentence.

Confounding factors in NLI \cite{wang2019if},  other natural language and machine learning-based applications have recently been studied as well. As pointed out by \cite{wang2019if}, statistical irregularities present in datasets result in higher-than-expected performance in video sentiment analysis \cite{DBLP:journals/corr/WangMMX16}, visual question answering \cite{DBLP:journals/corr/GoyalKSBP16} and medical applications \cite{yue2018deep}.
\subsection{Hypothesis Only Baselines}
Apart from proposing and building the hypothesis-only baseline model, Poliak et al. \cite{poliak_hypothesis_2018} tested it on the different NLI datasets which were categorized on the basis of their construction: (1) human-elicited (hypothesis and labels generated by humans); (2) human judged (hypothesis generated automatically with human annotated labels); and  (3) automatically recast datasets (minimal human intervention). Important to this paper are the recast datasets, where hypothesis-premise pairs were generated automatically using heuristics from the tasks they were constructed from. Specific examples of typical recast datasets are: Semantic Proto-Roles (SPR), Definite Pronoun Resolution (DPR) and FrameNet Plus (FN+). These are further expanded on in the next section.

The hypothesis-only baselines proved to be strong benchmarks to compare against newly models due to their performance on each of the NLI datasets. Their strong performance indicated that they might be "learning" patterns in the datasets that are external to true NLI. The authors specifically analyzed the presence of specific words, grammaticality, and lexical semantics.  The analysis of the conditional probability of the occurrence of specific labels in SNLI datasets given specific word revealed that such human elicited datasets may be biased to many label-specific terms if the data are not properly controlled. From the analysis of the grammaticality on the recast FrameNet Plus dataset, the authors found positive correlation between grammaticality and NLI labels --- higher grammaticality hypotheses tend to be labeled as ENTAILED. From the analysis of lexical semantics, a relatively high accuracy was found in property-driven hypotheses in the SPR dataset. Collectively, these findings suggest positively that patterns external to true language understanding do exist in currently used NLI datasets.

\subsection{Recast Semantic Proto-Role Dataset}
The idea of ``recasting semantic resources into a unified evaluation framework'' was proposed by White et al. \cite{white-etal-2017-inference}. The authors described the FraCaS dataset\cite{cooper-1996-fracas}, which was constructed based on semantic fine-grained probes.
This was followed by SNLI dataset, constructed with human elicited text-hypothesis pairs. In contrast, datasets such as FraCaS would not be properly designed for NLI task that requires large-scale data-driven computational semantics, for it is not practical for experts to author enough examples. The results of applying recasting strategy on three datasets -- SPR \cite{reisinger_semantic_2015}, FN+\cite{pavlick-etal-2015-framenet} and DPR\cite{rahman-ng-2012-resolving} -- suggested that a general approach of ``converting semantic classification examples to annotated textual inference pairs'' to train RTE models can be used as an automatic way that can remove human in the data generation loop.
DPR is a dataset that was originally constructed to resolve complex case of definite pronoun\cite{rahman-ng-2012-resolving}. White et al. \cite{white-etal-2017-inference} replaced the pronoun of each sentence with its correct and incorrect referent to produce hypothesis from DPR dataset. FN+ is an expanded version of FrameNet, constructed using crowdsourcing to filter out incorrect paraphrases manually \cite{pavlick-etal-2015-framenet}. The recast version of FN+ was further edited with crowdsourced judgements to generate entailed and non-entailed hypotheses.    

The original SPR dataset decomposes semantic roles into 16 finer-grained properties, shown in \textsc{table}~\ref{tab:16}. For example, given the role property ``instigation'', the system looks for whether a predicate's argument likely caused a given predicated situation. For purposes of recasting the dataset into an NLI form, judges were provided with the question ``How likely or unlikely each property was to hold of the argument in the context of the predicate'' and were asked to estimate the likelihoods on a five-point scale from 1 (very unlikely) to 5 (very likely). The scale was used to categorize the predictions, with a score from 1-3 resulting in ``not entailed'' label and a score of 4-5 would result in``entailed'' label, the ``neutral'' label was not considered in this dataset.

\begin{table}[h]
\centering
\label{tab:16}
\caption{Questions posed to SPR annotators\cite{white-etal-2017-inference}}
\begin{tabular}{|l|l|}
\hline
\textbf{Role property} & \textbf{How likely or unlikely is it that...}                                                                             \\ \hline
instigation            & ARG caused the PRED to happen?                                                                                            \\ \hline
volition               & ARG chose to be involved in the PRED?                                                                                     \\ \hline
awareness              & \begin{tabular}[c]{@{}l@{}}ARG was/were aware of being involved in \\ the PRED?\end{tabular}                              \\ \hline
sentient               & ARG was/were sentient?                                                                                                    \\ \hline
change of location     & ARG changed location during the PRED?                                                                                     \\ \hline
exists as physical     & ARG existed as a physical object?                                                                                         \\ \hline
existed before         & ARG existed before the PRED began?                                                                                        \\ \hline
existed during         & ARG existed during the PRED?                                                                                              \\ \hline
existed after          & ARG existed after the PRED stopped?                                                                                       \\ \hline
change of possession   & ARG changed possession during the PRED?                                                                                   \\ \hline
change of state        & \begin{tabular}[c]{@{}l@{}}ARG was/were altered or somehow changed \\ during or by the end of the PRED?\end{tabular}      \\ \hline
stationary             & ARG was/were stationary during the PRED?                                                                                  \\ \hline
location of event      & ARG described the location of the PRED?                                                                                   \\ \hline
physical contact       & \begin{tabular}[c]{@{}l@{}}ARG made physical contact with someone or \\ something else involved in the PRED?\end{tabular} \\ \hline
was used               & ARG was/were used in carrying out the PRED?                                                                               \\ \hline
pred changed arg       & The PRED caused a change in ARG                                                                                           \\ \hline
\end{tabular}
\end{table}

\section{Methodology and Experiment}
Hypothesis baselines trained on the SPR dataset
\footnote{The SPR data can be found at https://github.com/decompositional-semantics-initiative/DNC/raw/master/inference\_is\_everything.zip} 
perform considerably better than majority baselines and were shown to be comparable with the state of the art performance which at the time was a modified InferSent\cite{Conneau-InferSent-17} method. The InferSent method uses a encoder based on BiLSTM architecture with different pooling strategies to obtain sentence representation for each input sentence and a multi-layer perceptron classifier to predict the NLI tag.
The fact that a model that only encodes the hypothesis performs as well as the state of the art emphasize issues in both NLI models, and NLI datasets --- is the model merely picking up statistical patterns in the dataset to show good results? To further explore whether statistical patterns are present in this dataset, we propose a method that consists of two types of analysis on the recast SPR datset in our preliminary research.

\subsection{Proposed Method}
This project aims to explore potential statistical irregularities that could bias the performance of NLI models.  We focus specifically on recast datasets. The driving idea is that, by analyzing an automatically generated datasets from external tasks with minimal human intervention, we expect to see more biases that may not be spotted in datasets with human intervention.

Since proto-roles in hypotheses are distributed over the two labels in the SPR dataset, we are interested in whether they are responsible for the increased performance of the hypothesis-only model, which would be quantified as the proto-role bias.

\vspace{2mm}
\subsubsection{Chi-Square Test}
To test the association between NLI labels and proto-roles, we use the chi-square test:

\begin{equation}
\chi^2 = \sum_{i=1}^n\frac{(O_i-E_i)^2}{E_i}
\label{chi-square}
\end{equation}\\
where\par
$\chi^2$ = Pearson's cumulative test statistic\par
$O_i$ = the number of observations in each proto-role\par
$E_i$ = the expected count in each proto-role\par
$n$ = the number of proto-roles \\
The chi-square test of independence examines whether the observed pattern between variables provides enough evidence to support the independence of two variables, and thus, is appropriate for the task. Our null hypothesis is that there is no association between NLI labels and proto-roles. We conduct the chi-square test at $\alpha = .05$ level. 

\vspace{2mm}
\subsubsection{Proto-Role bias}
We determine the proto-role bias, or the majority baseline for each proto-role by the conditional probability of the majority label $l$ given the proto-role $pr$. A predictive bias would be captured if a specific majority baseline is significantly high. Poliak\cite{poliak_hypothesis_2018} described such words as ``give-away'' words that should be removed for a uniform distribution of the majority baseline across labels. These words have a great potential to affect the overall accuracy on NLI. The majority baseline for a given $pr$, $\text{maj}_{pr}$ is:
\begin{equation}
\label{eq:2}
\text{maj}_{pr}=\frac{\text{count}(l, pr)}{\text{count}(pr)},
\end{equation} where count($pr, l$) describes the number of sentences that are majority-labelled and contain the proto-role and count($pr$) is the number of sentences containing that proto-role.

For instance, sentence, \textit{The increasing caused a change in 14 members} will be counted as an instance for the proto-role ``instigation''. If there are 60 such hypotheses labelled ``entailed'' out of 100 in total, the proto-role bias for the ``instigation" proto-role would be 0.6.

\vspace{2mm}
\subsubsection{Proto-role Bias Calculation} 
We compute the overall proto-role bias in the dataset by summing the majority baselines for every $pr$ (for $N$ proto-roles): 
\begin{equation}
\label{eq:3}
\text{Overall Proto-role Bias}=\frac{\sum_{pr}^{N}\text{maj}_{pr}\times\text{count}(pr)}{\text{count}(total)}
\end{equation}

Given 100 instances of the proto-role ``instigation" and 100 instances of the proto-role ``awareness", if 110 (40 I, and 70 A) instances are labelled as ``non-entailed", then the overall majority baseline accuracy would be $\frac{110}{200} = .55$, however, if we consider the proto-role bias described in equations \eqref{eq:2} and \eqref{eq:3}, then the proto-role bias for each individual proto-role would be 0.6 (for ``instigation"), and 0.7 (for ``awareness"), giving us the overall proto-role bias (and equivalently, the accuracy) of
$\frac{0.6 \times 100 + 0.7 \times 100}{200}=0.65$.

When controlled for biases within individual proto-roles, we get improved accuracy measures, which is indicative of potential artifacts in the dataset that such hypothesis only models can pick up on, leading to an inflated sense of performance of the model.

While the above analysis only investigates the proto-roles in the hypotheses, There could be lexical biases due to many other such ``give-away" words. For both ``entailed'' and ``not-entailed'' labels, we rank words in their hypothesis sentences according to their frequency within each label and report the top 10 most frequent words. By applying the same conditional probability function, we can explore other highly biased lexical irregularities other than specific proto-roles.

\begin{table}[h]
\centering
\caption{SPR Dataset Statistics}
\label{tab:data-stat}
\begin{tabular}{|c|c|c|c|}
\hline
\multicolumn{1}{|l|}{} & \multicolumn{1}{c|}{\textbf{Train}} & \multicolumn{1}{c|}{\textbf{Dev}} & \multicolumn{1}{c|}{\textbf{Test}} \\ \hline
\textbf{Entailed}        & 43,148                               & 5,313                              & 5,341                               \\ \hline
\textbf{Not-entailed}    & 80,707                               & 9,983                              & 10,115                              \\ \hline
\textbf{Total}         & 123,855                              & 15,296                             & 15,456                              \\ \hline
\end{tabular}
\end{table}

\subsection{Dataset Description}
The SPR dataset is split into train, dev and test sets, the statistics of which are presented in \textsc{table~\ref{tab:data-stat}}. The two NLI labels used in this dataset are ``entailed'' and ``not-entailed''. As is the case with several previous NLI datasets \cite{bowman_large_2015, williams2018broad}, the assumption that both premise and hypothesis refer to the same single event holds here as well. In this dataset, the ``neutral'' relation is not considered.

\section{Results and Discussion}
We performed chi-square test for independence between NLI labels and proto-roles on each of the train, dev, and test sets in SPR. From \textsc{table}~\ref{tab:chi}, we find no independence between NLI labels and proto-roles ($p < .05$ for train, dev, and test sets). It is likely that the occurrence of a proto-role would impact the prediction of the relation label if the frequency of each proto-role in the dataset is not uniformly distributed across the relation labels.

\begin{table}[h]
\centering
\caption{Results of Chi Square Test}
\label{tab:chi}
\begin{tabular}{|c|c|c|c|}
\hline
\multicolumn{1}{|c|}{\textbf{}} & \textbf{Train} & \multicolumn{1}{c|}{\textbf{Dev}} & \textbf{Test}  \\ \hline
\textbf{$\bm{\chi^2}$} & 30,632 & 3897.1 & 3,781.1 \\ \hline
\textbf{df}            & 15     & 15     & 15      \\ \hline
\textbf{\textit{p}}       & \multicolumn{1}{l|}{\textless{} 2.2e-16} & \textless{} 2.2e-16                & \multicolumn{1}{l|}{\textless{} 2.2e-16} \\ \hline
\end{tabular}
\end{table}

The results in \textsc{table}~\ref{tab:overall} depicts a general picture comparing the proto-role bias to the majority baseline. Here, we compute the accuracy for a model that follows from our proto-role bias quantification, i.e., the model predicts the relation label that occurs the most within instances of that proto-role (calculated from the training set). We call this model a ``Proto-Role Biased Model" (PRBM). We find the accuracy of PRBM to be higher than the majority baseline in all three splits, indicating that a model can get a higher score than predicting the majority label by simply predicting irregularities within proto-roles. 

\begin{table}[h]
\centering
\caption{Overall Majority Baseline and Proto-Role Bias}
\label{tab:overall}
\begin{tabular}{|c|c|c|c|}
\hline
\multicolumn{1}{|c|}{\textbf{}} & \multicolumn{1}{c|}{\textbf{Train}} & \multicolumn{1}{c|}{\textbf{Dev}} & \multicolumn{1}{c|}{\textbf{Test}} \\ \hline
\textbf{MAJ}                    & 0.6635                              & 0.6527                            & 0.6544                             \\ \hline
\textbf{PRBM}                & 0.7460                              & 0.7492                            & 0.7473                             \\ \hline
\end{tabular}
\end{table}

Upon further inspection, we see that the performance of PRBM on the SPR dataset is not consistent across different proto-roles. \textsc{table}~\ref{tab:pr} indicates that there are highly biased proto-roles such as ``stationary'' that gives a score of 0.96, which suggests the existence of certain proto-roles that are heavily indicative of a particular entailment level. 

\begin{table}[h]
\centering
\caption{Proto-Role Bias}
\label{tab:pr}
\begin{tabular}{|c|c|c|c|}
\hline
\multicolumn{1}{|c|}{\textbf{Proto-Role Property}} & \textbf{Train} & \textbf{Dev} & \textbf{Test} \\ \hline
Instigation                               & 0.6327         & 0.6308       & 0.6335        \\ \hline
Volition                                  & 0.6463         & 0.6496       & 0.6449        \\ \hline
Awareness                                 & 0.6108         & 0.5994       & 0.6087        \\ \hline
Sentient                                  & 0.7618         & 0.7626       & 0.7547        \\ \hline
Change of location                        & 0.9292         & 0.9320       & 0.9327        \\ \hline
Exists as physical                        & 0.6583         & 0.6538       & 0.6594        \\ \hline
Existed before                            & 0.6562         & 0.6590       & 0.6480        \\ \hline
Existed during                            & 0.8601         & 0.8567       & 0.8468        \\ \hline
Existed after                             & 0.6999         & 0.7291       & 0.6967        \\ \hline
Change of possession                      & 0.9340         & 0.9446       & 0.9389        \\ \hline
Change of state                           & 0.6375         & 0.6485       & 0.6522        \\ \hline
Stationary                                & 0.9631         & 0.9634       & 0.9627        \\ \hline
Location of event                         & 0.9200         & 0.9121       & 0.9172        \\ \hline
Physical contact                          & 0.8544         & 0.8692       & 0.8582        \\ \hline
Was used                                  & 0.5391         & 0.5272       & 0.4482        \\ \hline
Pred changed arg                          & 0.6333         & 0.6485       & 0.6511        \\ \hline
\end{tabular}
\end{table}

Besides proto-roles biases, our research reveals lexical biases toward non-entailment labels. We present the top 5 frequent words in the dev split of the dataset in \textsc{table}~\ref{tab:lexical}. The probabilities of ``not-entailed'' label given the top 5 frequent words illustrate that those frequent words could lead to a higher accuracy on predicting ``not-entailed'' label. 

\begin{table}[H]
\centering
\caption{Lexical Analysis for Not-Entailed Cases}
\label{tab:lexical}
\begin{tabular}{|c|c|c|}
\hline
\multicolumn{1}{|c|}{\textbf{Word}} & \textbf{P(l$|$w)} & \textbf{Freq} \\ \hline
market                     & 0.7326          & 211           \\ \hline
that                       & 0.8189          & 208           \\ \hline
stock                      & 0.6612          & 201           \\ \hline
company                    & 0.6471          & 176           \\ \hline
they                       & 0.6324          & 172           \\ \hline
\end{tabular}
\end{table}

In contrast, words are more uniformly distributed in hypotheses that are labelled ``entailed''. As illustrated in \textsc{table}~\ref{tab:lexical-entail}, the top frequent words ``stock'', ``they'', ``company'', ``some'' and ``making'' all appear in the hypotheses for less than 50 percent of time. However, it is not conclusive to state that there is no observed word that could bias the ``entailed''  prediction. As there are three out of five words that are overlapped in both tables, it is reasonable to speculate that the pattern does exist to predict lexical irregularities in low frequency. 

\begin{table}[H]
\centering
\caption{Lexical Analysis for Entailed Cases}
\label{tab:lexical-entail}
\begin{tabular}{|c|c|c|}
\hline
\multicolumn{1}{|c|}{\textbf{Word}} & \textbf{P(l$|$w)} & \textbf{Freq} \\ \hline
stock                     & 0.3388          & 103           \\ \hline
they                       & 0.3676          & 100           \\ \hline
company                      & 0.3529          & 96           \\ \hline
some                    & 0.4178          & 94           \\ \hline
making                       & 0.3571          & 90           \\ \hline
\end{tabular}
\end{table}

\section{Conclusion and Future Work}

The task of NLI requires understanding and representing semantic and commonsense knowledge in a manner that allows inference between pairs of events. Existing models for NLI, e.g. neural networks, show desirably high performance on such tasks but the extent to which they actually utilize semantic knowledge as opposed to pattern matching is unclear. This work contributes to the growing literature of diagnosing model performance by highlighting certain specific patterns that might be ``learned" by a model, which have nothing to do with semantic knowledge.
Specifically, we built on the hypothesis-only baseline method proposed by \cite{poliak_hypothesis_2018} to diagnose word-level (lexical) biases in an existing NLI dataset (SPR).
Our experiments show that the presence of statistical irregularities, due to the differences in distribution of entailment labels over proto-roles, could allow NLI models to show greater performance due to factors external to what is required to perform inferences over texts. This is accomplished by first showing non-zero dependence between proto-role frequences and entailment labels, and then measuring performance by controlling for proto-roles. 
Our analysis is further augmented by showing other lexical irregularities that could contribute to NLI models' performance. We emphasize that such patterns can lead the model to predict the true label based on the corresponding lexical features on the basis of frequency instead of testing true semantic knowledge.

We hope our findings could provide an insight for NLI datasets developers and users to avoid exploitable irregularities. In the future, we will explore biases in such datasets to understand factors that might contribute to the models’ performance. A deeper dive into whether or not existing models actually exploit such lexical biases would be a worthwhile endeavor to pursue. To do so will involve examining model output based on biased and unbiased/debiased examples of NLI instances.

\bibliographystyle{./bibliography/IEEEtran}
\bibliography{./bibliography/citations}

\end{document}